\documentclass[conference]{IEEEtran}

\usepackage{times}
\usepackage{epsfig}
\usepackage{graphicx}
\usepackage{cite}
\usepackage{amsmath,amssymb,amsfonts}
\usepackage{algorithmic}
\usepackage{graphicx}
\usepackage{textcomp}
\usepackage{xcolor}

\usepackage[pagebackref=true,breaklinks=true,letterpaper=true,colorlinks,bookmarks=false]{hyperref}

\begin{document}

\title{Applications of the Streaming Networks}

\author{\IEEEauthorblockN{Sergey Tarasenko\IEEEauthorrefmark{1},
Fumihiko Takahashi\IEEEauthorrefmark{2}}

\IEEEauthorblockA{Next-Gen Mobility Division, Mobility R$\&$D Group,
JapanTaxi\\
Tokyo, Japan\\
Email: \IEEEauthorrefmark{1}tarasenko.sergey@japantaxi.co.jp,
\IEEEauthorrefmark{2}fumihiko.takahashi@japantaxi.co.jp}}

\maketitle

\begin{abstract}
Most recently Streaming Networks (STnets) have been introduced as a mechanism of robust noise-corrupted images classification. STnets is a family of convolutional neural networks, which consists of multiple neural networks (streams), which have different inputs and their outputs are concatenated and fed into a single joint classifier. The original paper has illustrated how STnets can successfully classify images from Cifar10, EuroSat and UCmerced datasets, when images were corrupted with various levels of random zero noise. In this paper, we demonstrate that STnets are capable of high accuracy classification of images corrupted with Gaussian noise, fog, snow, etc. (Cifar10 corrupted dataset) and low light images (subset of Carvana dataset). We also introduce a new type of STnets called Hybrid STnets. Thus, we illustrate that STnets is a universal tool of image classification when original training dataset is corrupted with noise or other transformations, which lead to information loss from original images.
\end{abstract}

\section{Introduction}

Recently, CNNs\footnote{Convolutional Neural Networks} with more than one processing stream have started to gain popularity. To our knowledge, the first two-stream CNN was introduced by Chorpa \cite{Chopra2005LearningAS} and it is widely known as a ``Siamese network". The motivation behind two streams is that each of the streams carries information about a dedicated image. Images fed to the streams are different.

Most recently, two-stream CNNs have been used for the vast variety of recognition, segmentation and classification tasks such as similarity assessment (Siamese networks and pseudo-Siamese \cite{Chopra2005LearningAS, Zagoruyko2015LearningTC}), change detection and classification \cite{Varghese2018ChangeNetAD}, action recognition in videos \cite{Simonyan2014TwoStreamCN}, one-shot image recognition \cite{Koch2015SiameseNN}, simultaneous detection and segmentation \cite{Hariharan2014SimultaneousDA}, human-object interaction recognition \cite{Gkioxari2017DetectingAR}, group activity recognition \cite{Azar2018AMC}, etc.

Finally, multi-stream CNN has been successfully employed for recognition of 2D projections of the 3D image \cite{su15mvcnn}.

Most recently, it has been illustrated that Streaming networks (STnets) \cite{Tarasenko2019StreamingNE} are capable of recognition of zero noise-corrupted images with moderate accuracy without using of special techniques or training data augmentation.

In this study, we investigate if the STnets are capable of dealing with other types of image distortion rather than a simple random zero noise. Furthermore, similar mechanism of pixel value zeroing is intrinsic for low light images. Therefore, we also investigate low light image classification performance of STnets. 

\section{Streaming Networks Architecture}

STnets are constructed of multiple parallel streams, whose outputs are concatenated and fed into a classifier. During training, parameters of all streams are tuned independently. Each stream is taking a certain piece of information and its input is different from the inputs of the other streams. 

In the original study by Tarasenko and Takahashi \cite{Tarasenko2019StreamingNE}, each stream had input in the form of the image intensity slice. It is possible, however, to use other ways to split original image into slices. For example, 
Pinckaers et al. \cite{Pinckaers2019StreamingCN} used images patches (spatial slice) as input into their Streaming Network.

Next, we explain the mathematical concepts behind Streaming Networks proposed by Tarasenko and Takahashi \cite{Tarasenko2019StreamingNE}.

Eq. (\ref{eq:basis_vectors}) introduces an image representation by means of the orthogonal vectors, which form an orthogonal basis of some image representation space:

\begin{equation}
Img^{*} = \sum^{n}_{i=1} \alpha_{i} v_{i}
\label{eq:basis_vectors}
\end{equation}

where $\alpha_{i}$ is a weight coefficient for $i$-th orthogonal basis vector $v_{i}$, $i$ = 1, ..., $n$, and $Img^*$ is an image representation.

\begin{figure}[t]
\begin{center}
\includegraphics[width=1.0\linewidth]{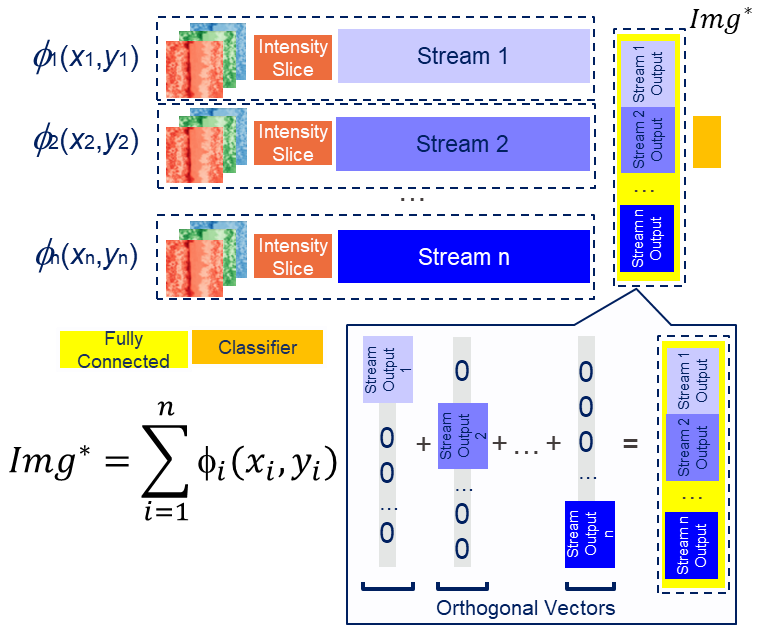}
\end{center}
\caption{Streaming networks as orthogonal basis.}
\label{fig:orthobasis}
\end{figure}

\begin{figure*}[t]
\begin{center}
\includegraphics[width=1.0\linewidth]{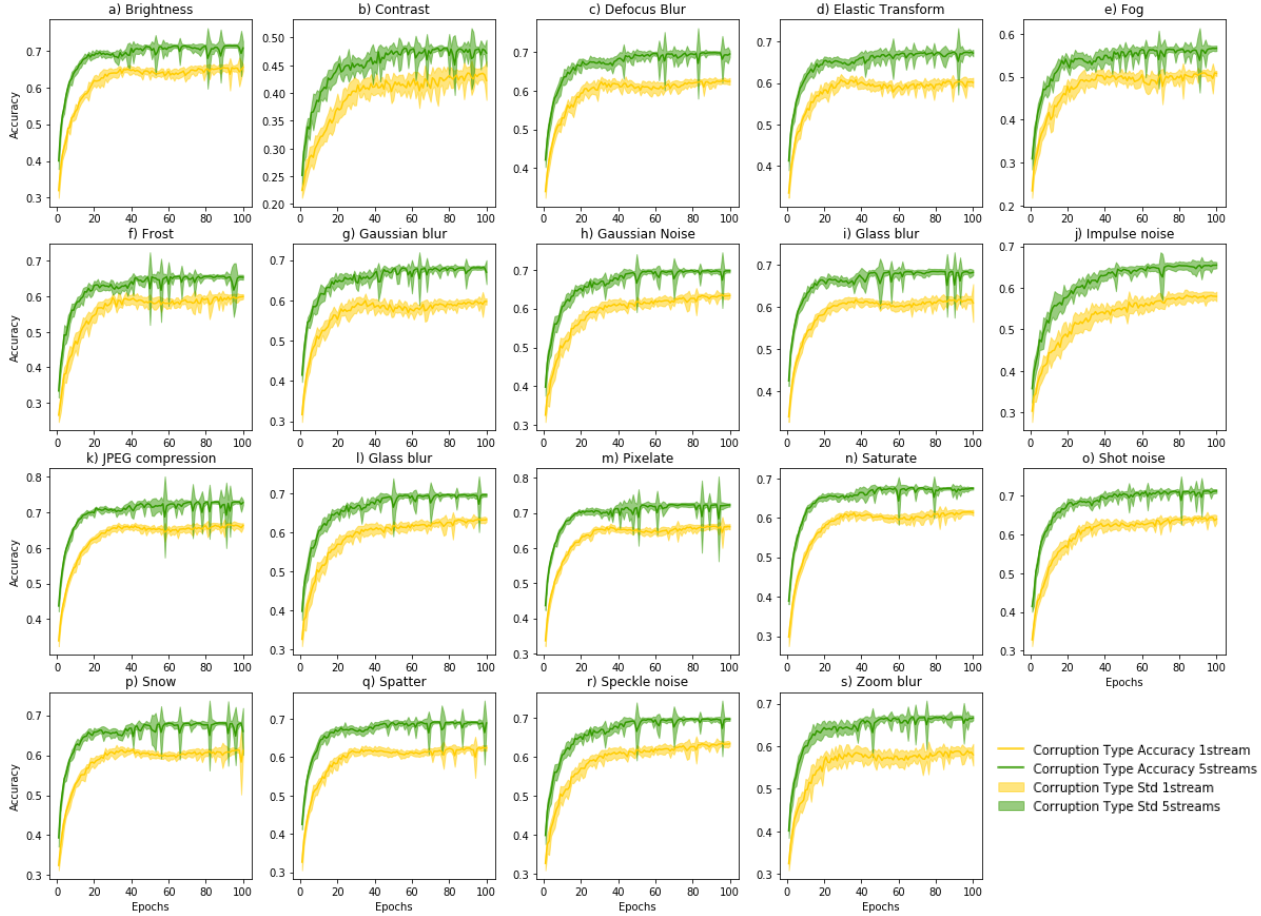}
\end{center}
\caption{Illustration of robust recognition of corrupted images from Cifar10 Corrupted dataset by Streaming Networks.}
\label{fig:cifar10_corr_res}
\end{figure*}

As the next step, we can generalize eq. (\ref{eq:basis_vectors}) by providing orthogonal basis functions $\phi_{i}(x_{i},y_{i}) $ instead of basis vectors $v_{i}$ \cite{Olshausen1996EmergenceOS}:

\begin{equation}
Img^{*} = \sum^{n}_{i=1} \alpha_{i}\phi_{i}(x_{i},y_{i})
\label{eq:basis_functions}
\end{equation}

where pair ($x_{i},y_{i}$), $i$ = 1, ..., $n$, represents some part of an original image.

If now, we set all weighting coefficients $\alpha_i$, $i$=1,...$n$, to 1, we will get eq. (\ref{eq:basis_functions_unit_weight}):

\begin{equation}
Img^{*} = \sum^{n}_{i=1} \phi_{i}(x_{i},y_{i})
\label{eq:basis_functions_unit_weight}
\end{equation}

Eq. \ref{eq:basis_functions_unit_weight} is the gist of the STnets.

\section{Experiments}
For all our experiments, we use Adam optimizer with learning rate of 0.0001 accompanied by $\beta$1 = 0.99, $\beta$2 = 0.9 and $\epsilon$ = 1e-08, and run all the trainings for 100 epochs. We use SoftMax classifier for all the experiments.

A 1-stream simple CNNs, which is employed as a single stream in some of experiments, has the following structure: 
1) conv layer with 32 7x7 filters plus ReLU activation and 2x2 Max-pooling layers;
2) conv layer with 64 5x5 filters plus ReLU activation and 2x2 Max-pooling layers;
3) conv layer with 128 3x3 filters plus ReLU activation and 2x2 Max-pooling layers;
4) conv layer with 256 1x1 filters plus ReLU activation and 2x2 Max-pooling layers;
5) conv layer with 10 1x1 filters plus ReLU activation and 2x2 Max-pooling layers;
6) fully connected layer;
7) SoftMax layer with the number of output neurons corresponding to the number of classes.

\subsection{Classification of Corrupted Images}
On this section, we extend previous experiments with random zero-noise with two new conditions. First, we test STnets using Cifar10 corrupted \footnote{https://zenodo.org/record/2535967} dataset. Second, using the random zero-noise, we test performance of three 5-streams STnet based on a simple CNN, VGG16 \cite{Simonyan2014VeryDC} and ResNet50 \cite{He2015DeepRL} models .

\textbf{Experiments with Cifar10 Corrupted dataset.}
Cifar10 corrupted dataset contains original cifar10 images corrupted with 19 types of distortions, where brightness change, contrast change, defocus blur, elastic transform, fog, frost, Gaussian blur, Gaussian noise, glass blur, impulse noise, JPEG compression noise, motion blur, pixelate, saturation, shot noise, snow, spatter, speckle noise and zoom blur. These corruptions are explained in details in this study \cite{Hendrycks2018BenchmarkingNN}. 

We test performance of the 1-stream simple CNN and 5-streams STnet using Cifar10 images as a training set and Cifar10 corrupted images as a test set. We run each network for 10 times. The results are presented in Fig. \ref{fig:cifar10_corr_res}.

The results indicate that STnets outperform 1-stream simple CNN for all types of image distortions. 

\textbf{Different types of the Streaming Networks.} 
Here we address the issue of whether STnets based on different types of CNNs behave differently under the same experimental conditions.

To evaluate performance of three different streaming networks, we have used of 10 runs of each networks for 10$\%$ of noise using cifar10 dataset. 

\begin{figure}[t]
\begin{center}
\includegraphics[width=0.8\linewidth]{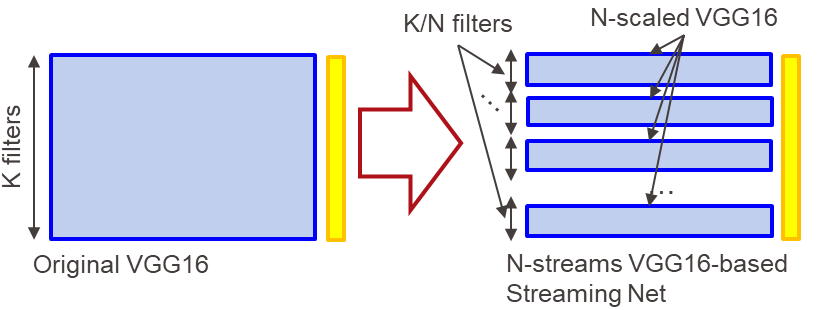}
\end{center}
\caption{N-scaled VGG16.}
\label{fig:nscaled}
\end{figure}

In the case of VGG16 network we use its scaled version when number of filters in each convolution layer is reduced by approximately a factor of $n$. A structure of n-scaled VGG16 network is presented in Fig. \ref{fig:nscaled}.

\begin{figure}[t]
\begin{center}
\includegraphics[width=1.0\linewidth]{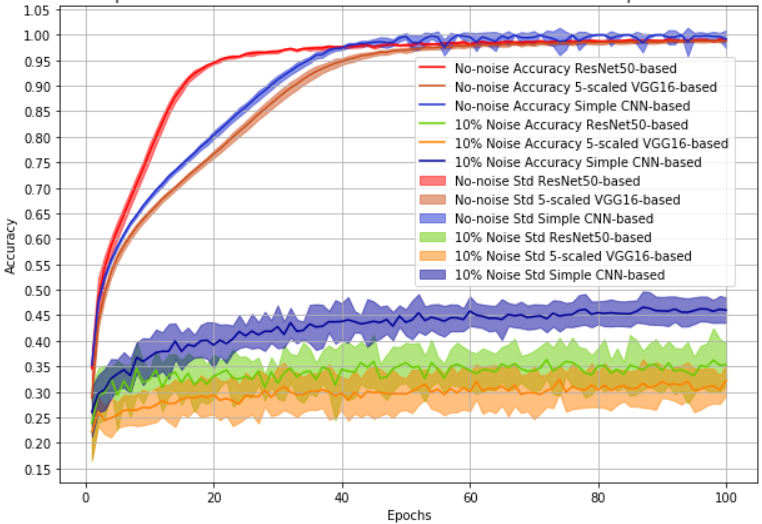}
\end{center}
\caption{Compare 5-streams Streaming Networks based on simple CNN, 5-scaled VGG16 and ResNet50 models.}
\label{fig:stream_models}
\end{figure}

The results are presented in Fig.\ref{fig:stream_models}: 1) simple convnet-based Streaming network exhibits the highest test accuracy for noise-corrupted images classification and the lowest variance; 2) ResNet50-based Streaming Network outperforms 5-scaled VGG16 in terms of test accuracy; 3) ResNet50-based network exhibits the fastest convergence of training accuracy to 100$\%$.

This experiment illustrates that there is indeed variability in performance of STnets depending on the CNNs used as streams.

\section{Classification of Low Light Images}
To test whether STnets are capable of recognition of low-light images after having been trained on normal light images, we employ Carvana\footnote{https://www.kaggle.com/c/carvana-image-masking-challenge} dataset. 

This dataset was originally designed for image segmentation tasks, however we use this dataset for classification task. For this purpose, we have selected randomly front view of 10 car models. A new sub-dataset consists of 2320 images. For our experiments we also resize original Carvana images to 128x191 pixels (height x width).

As a single stream network, we employed original full scale VGG16 network. As a Streaming Network we used 6-stream STnet network with one stream based on 5-scaled VGG16 and remaining five streams based a simple CNN. We call STnets containing different types of CNNs as streams to be \textit{hybrid STnets}. 

To imitate low light condition, we have applied Gamma transformation with $\gamma$=5.8:

\begin{equation}
Img^{*} = (Img/255)^{\gamma}*255
\label{eq:gamma _transform}
\end{equation}

\begin{figure}[t]
\begin{center}
\includegraphics[width=1.0\linewidth]{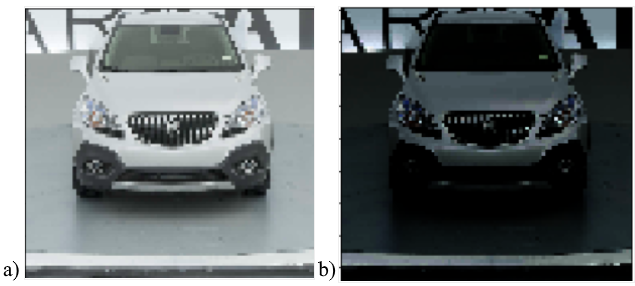}
\end{center}
\caption{Apply gamma transformation: a) original clipped image from Carvana dataset; 2) output of Gamma transformation with $\gamma$=5.8.}
\label{fig:gamma}
\end{figure}

Examples of original lighting condition and low-light after gamma transformation are presented in Fig. \ref{fig:gamma}. To illustrate the effect of Gamma transformation we have clipped original image to focus only on the area with car front view.

Throughout experiments, we split our data into 2000 train samples and 320 test sample. All test samples undergo gamma transformation. Each network was run for 10 times.

\begin{figure}[t]
\begin{center}
\includegraphics[width=1.0\linewidth]{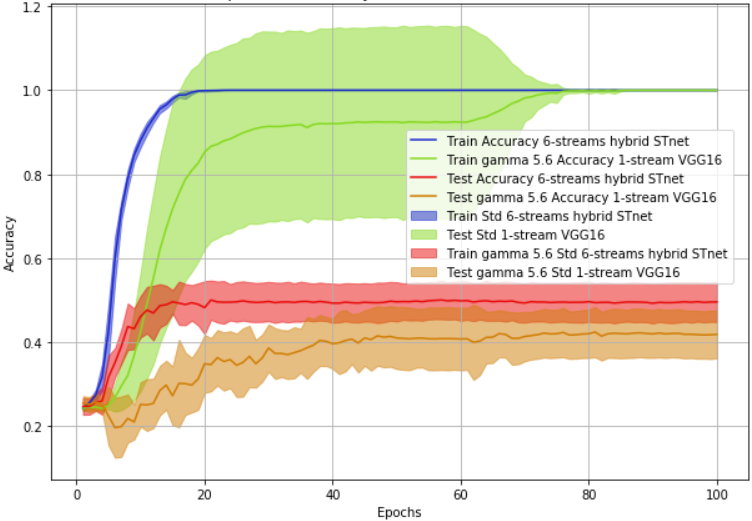}
\end{center}
\caption{Low light image recognition with 1-stream VGG16 vs. 5-streams 5-scaled VGG16 Streaming Network.}
\label{fig:low_light}
\end{figure}

The results are presented in Fig. \ref{fig:low_light}. One can infer that in the case of VGG16 network train accuracy is converging to 100$\%$ slower and with much higher variance than in the case of hybrid STnet. On the other hand, hybrid STnet exhibits significantly higher accuracy of low-light images (test accuracy) classifications with lower variance than in the case of VGG16 network.

Therefore we illustrate that STnets, trained on images using normal lighting conditions, are capable of moderate accuracy recognition of the low-light images.

\section{Discussion and Conclusion}

In this paper, we have extended a number of applications of STnets to the use cases of images corrupted by various distortions and low light conditions. In both cases, STnets have outperformed 1-stream neural networks in terms of recognition accuracy of test data, i.e., corrupted and low light images. Therefore, it is possible to enable robust recognition of corrupted and low light images simply by applying STnets.

We have also considered various types of CNNs as a single stream for the STnets. We have found that STnets constructed from simple CNN model outperforms the ones constructed either from 5-scaled VGG16 or ResNet50 networks.

To conclude, we emphasize that STnets is a new architecture, which enables robust recognition of corrupted images under variety of distortion types. We hope this result will contribute to robust object recognition from corrupted surveillance images, which are subject to corruptions due to lighting or weather condition, speed, distance and camera angle.

{\small

}

\end{document}